\documentclass[sigconf]{acmart}
\AtBeginDocument{%
 }
\usepackage{adjustbox,bbm,multirow,multicol}
\usepackage{natbib}

\setcopyright{acmcopyright}
\copyrightyear{2023}
\acmYear{2023}
\acmDOI{}

\acmConference[MM '23] {Proceedings of the 31st ACM International Conference on Multimedia}{October 29--November 3, 2023}{Ottawa, ON, Canada.}
\acmBooktitle{Proceedings of the 31st ACM International Conference on Multimedia (MM '23), October 29--November 3, 2023, Ottawa, ON, Canada}

\acmPrice{15.00}
\acmISBN{}

\begin{document}

\title{Learning Implicit Entity-object Relations by Bidirectional Generative Alignment for Multimodal NER}

\author{Feng Chen}
\authornote{Feng Chen and Jiajia Liu contributed equally to this work and should be considered co-first authors.}
\affiliation{%
  \institution{Ant Group}
  \country{}}
\email{chenfeng1271@gmail.com}

\author{Jiajia Liu}
\affiliation{%
  \institution{Ant Group}
  \country{}}
\email{lekun.ljj@antgroup.com}

\author{Kaixiang Ji}
\affiliation{%
  \institution{Ant Group}
  \country{}}
\email{kaixiang.jkx@antgroup.com}

\author{Wang Ren}
\affiliation{%
  \institution{Ant Group}
  \country{}}
\email{renwang.rw@antgroup.com}

\author{Jian Wang}
\authornote{Jian Wang is the corresponding author.}
\affiliation{%
  \institution{Ant Group}
  \country{}}
\email{bobblair.wj@antgroup.com}

\author{Jingdong Chen}
\affiliation{%
  \institution{Ant Group}
  \country{}}
\email{ jingdongchen.cjd@antgroup.com}



\begin{abstract}
The challenge posed by multimodal named entity recognition (MNER) is mainly two-fold: (1) bridging the semantic gap between text and image and (2) matching the entity with its associated object in image. Existing methods fail to capture the implicit entity-object relations, due to the lack of corresponding annotation. In this paper, we propose a bidirectional generative alignment method named BGA-MNER to tackle these issues. Our BGA-MNER consists of \texttt{image2text} and \texttt{text2image} generation with respect to entity-salient content in two modalities. It jointly optimizes the bidirectional reconstruction objectives, leading to aligning the implicit entity-object relations under such direct and powerful constraints. Furthermore, image-text pairs usually contain unmatched components which are noisy for generation. A stage-refined context sampler is proposed to extract the matched cross-modal content for generation. Extensive experiments on two benchmarks demonstrate that our method achieves state-of-the-art performance without image input during inference.
\end{abstract}




\maketitle

\section{Introduction}

\begin{figure}[t]
\includegraphics[width=1\linewidth]{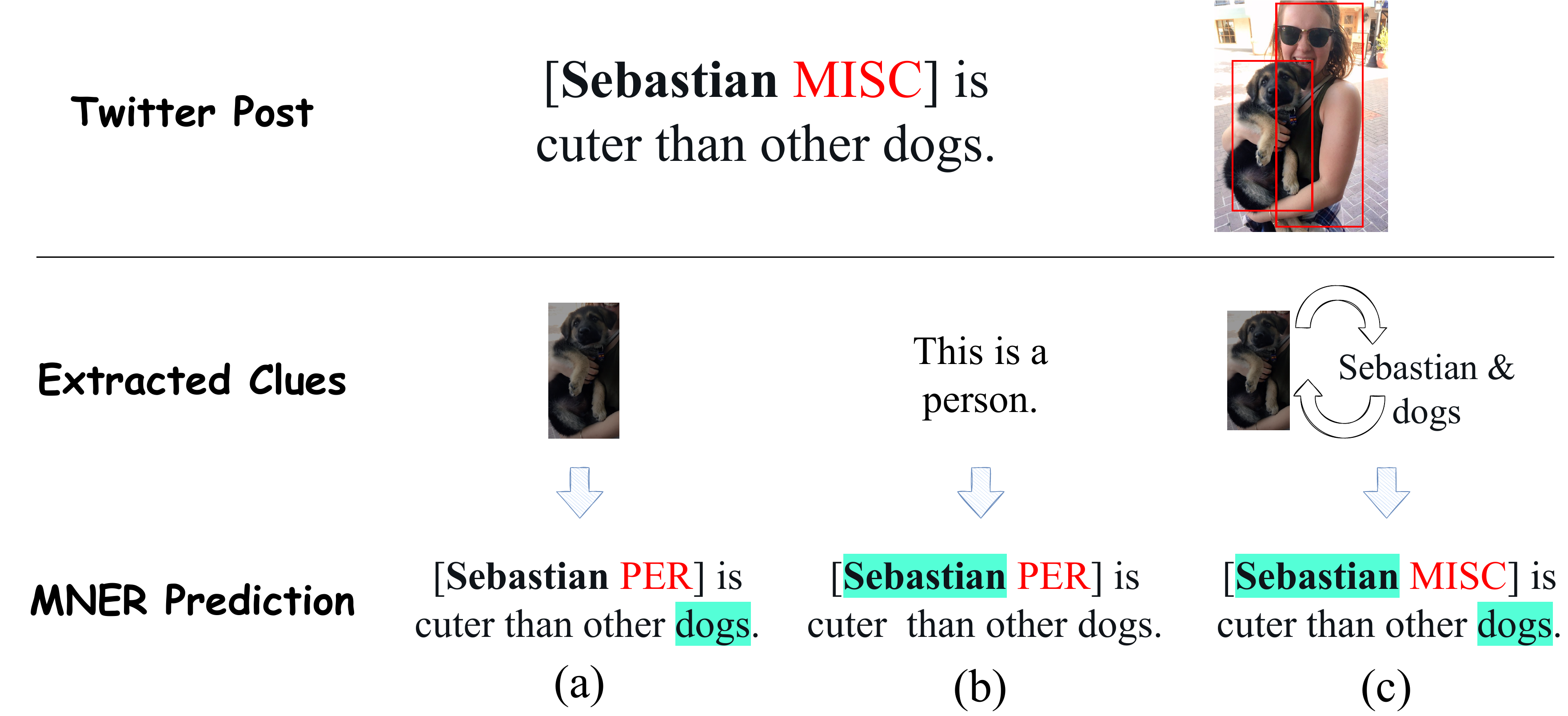}
\caption{The comparison of existing approaches in cross-modal alignment: (a) attention-based and visual grounding-based approaches, (b) knowledge-prompt-based approaches, (c) our BGA-MNER. The light green in font background denotes the textual correspondence of extracted visual clues.}
\label{fig1}
\end{figure}

Multimodal named entity recognition (MNER) \cite{uamner,umgf,zhao2021modeling} aims to extract predefined named entities from unstructured text with visual clues from the relevant images. The auxiliary visual clues are expected to improve the recognition of ambiguous multisense or out-of-vocabulary words. Compared to other vision-language tasks, e.g., image-text retrieval  \cite{xpool} and referring expression comprehension \cite{mattnet}, MNER is a more challenging task, due to widely-concerned semantic gap between image-text modality, and difficulties in aligning implicit entity-object relations without corresponding annotation. As illustrated in Figure \ref{fig1}, apart from the cross-modal semantic gap, this sample contains word `Sebastian', which can refer to either a person or a pet without further contextual information. Different from the obvious correlation between word `dog' and dog image, the relation between the entity `Sebastian' and dog is implicit in multimodal semantic context \cite{cat}.

 Previous works of MNER manage to fully exploit the helpful visual information and suppress the interference, which can be roughly divided into two categories. One line of the efforts focuses on utilizing entity-object relation with either cross-modal attention \cite{gcn,hvpnet} or visual grounding technology \cite{fmit,umgf}. 
 Another line adopts external knowledge, such as machine reading comprehension \cite{mrc}, knowledge prompt \cite{prompt} and expanded label words \cite{cat}, to induce knowledge-image or knowledge-entity co-occurrence. 

Despite their success, existing methods fail to directly capture the implicit entity-object relations \cite{cat,fmit}. As shown in Figure \ref{fig1} (a), attention and visual grounding-based methods tend to align concrete dog in text and image but ignore the desired `Sebastian', leading to recognizing `Sebastian' as PER type depending on textual representation. As for the knowledge-based methods shown in Figure \ref{fig1} (b), the top related prompt after querying the image is about person which is the main object in image. However, such information misleads the model that `Sebastian' is a person, so as to classify it as PER type. The crux of the issue lies in two aspects. (1) Previous methods are learned under text-dominated NER annotations. It is insufficient for enhancing cross-modal alignment \cite{cat}, causing heavy modality bias on textual information and misalignment of entity-object relations. (2) The potential entities in the sentence are often presented with names, which are harder than simple nouns for cross-modal alignment. Therefore, existing models may focus more on nouns-object relations while and the entity-object relations are still under-explored \cite{prompt}. 

\begin{figure*}[h]
\centering
\includegraphics[width=0.9\linewidth]{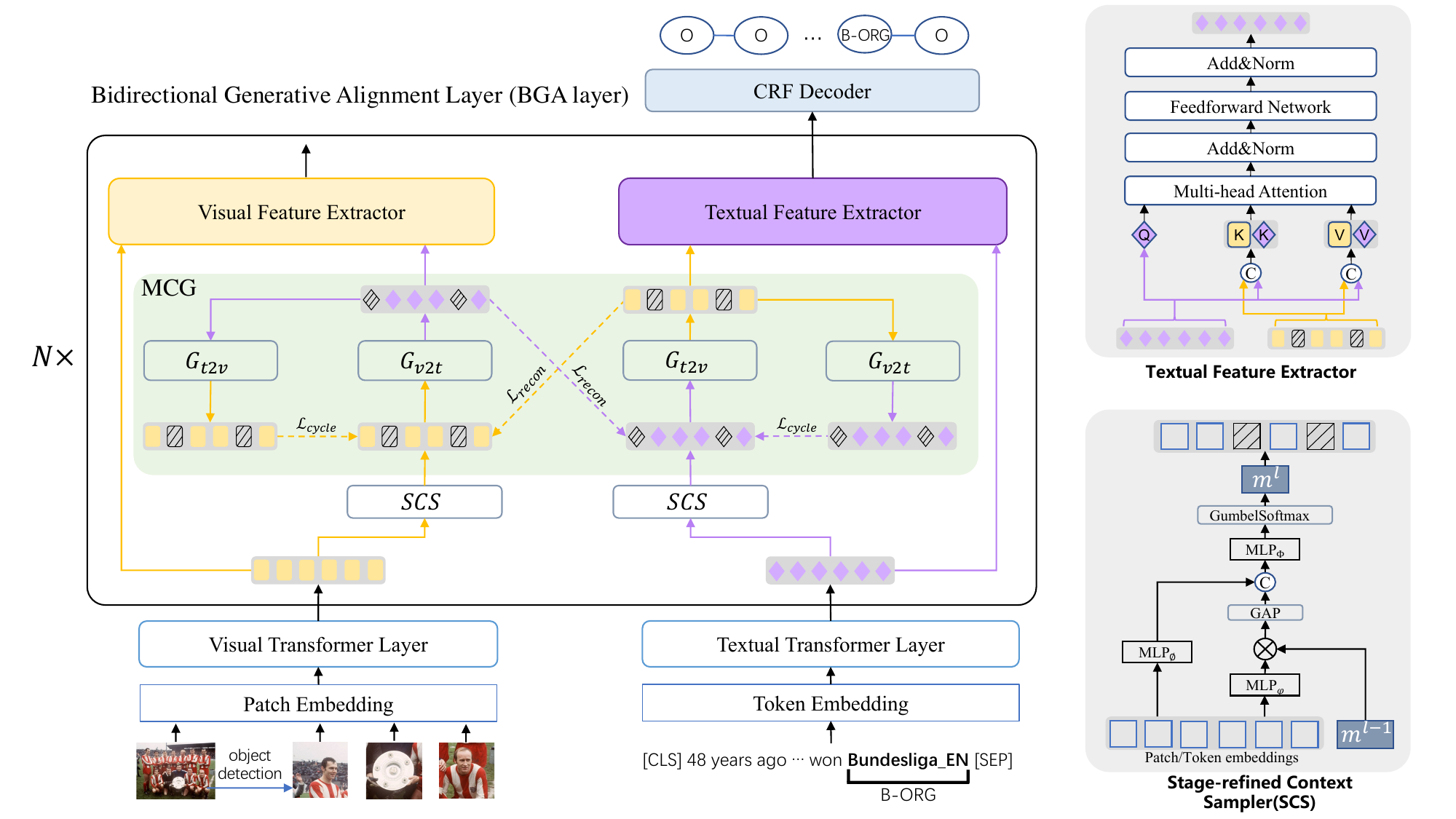}
\caption{Overview of our BGA-MNER. Our bidirectional generative alignment layer (BGA layer) consists of two Stage-refined Context Samplers (SCS), a Multi-level Cross-modal Generator (MCG) and two feature extractors. During inference, we only process the textual branch without cycle generation.}
\label{overview}
\end{figure*}

In this paper, we propose a novel bidirectional generative alignment method named BGA-MNER to address the above issues. The main idea of BGA-MNER is to use matched two-modality content for latent image2text and text2image generation. Such cross-modal generation supervision focuses on cross-modal alignment and alleviates the modality bias caused by NER supervision. Besides, by involving the entity and its corresponding object in bidirectional cross-modal generation, our method aligns the implicit entity-object relations under a direct and powerful constraint. For example, by enforcing  the entity `Sebastian' to generate the dog object and vice versa in Figure \ref{fig1}(c), our model learns the entity-object relation mapping directly during training and generates the desired entity-aligned visual features according to given entity during inference. Thus, we call this generation as generative alignment. To be specific, first, we propose a Stage-refined Context Sampler (SCS) to extract the most relevant content in image-text pairs by pruning unmatched tokens/patches. 
As shown in Figure \ref{fig1} (c), SCS samples the word `Sebastian'\&`dogs' and removes the irrelevant content. Second, we design a Multi-level Cross-modal Generator (MCG) to generate corresponding content of one modality with sampled content of the other modality, e.g., generating the visual dog content with given `Sebastian'\&`dogs' and vice versa. This bidirectional generation directly learns the implicit relations between `Sebastian' and dog with supervision. To ensure the successful mutual translation, we further take advantage of cycle consistency from CycleGAN \cite{cycle}, (e.g., the generated visual dog content is expected to  back-generated to `Sebastian'\&`dogs' in Figure \ref{fig1} (c)), to avoid under-constraint cross-modal mapping. With the help of aforementioned modules, the alignment of implicit entity-object relations is consolidated . Our contributions could be summarized as follows: 




\begin{itemize}
\item \textbf{A novel framework for MNER.} We propose a new end-to-end Transformer MNER framework, which aligns the two modal features in a generative manner. To the best of our knowledge, this is the first attempt to introduce bidirectional generative alignment for MNER.
\item \textbf{Image-free merit during inference.} By replacing the real visual feature with the generated one for cross-modal interaction, our framework is practical in dealing with text-only inputs and robust to noises from images.
\item \textbf{State-of-the-art performance.} Extensive experiments on Twitter2015 and Twitter2017 benchmarks indicate that our method outperforms existing state-of-the-art methods. Our model also shows superiority of aspects in cross-domain generalization and modality-missing robustness.
\end{itemize}

\section{Related Work}

\textbf{Multimodal NER.} 
Existing MNER methods mainly exploit the effective visual information and suppress the interference information \cite{riva,srpbert,umt, zhang2023aligning}. For attention-based works, MAF \cite{maf} proposes a cross-modal co-attention module to calculate the similarity score between text and image, then uses it as the proportion of visual information for cross-modal fusion. To solve the imprecise and biased cross-modal correspondence in attention-based methods, FMIT \cite{fmit} extends the lattice structure from Chinese NER \cite{flat} to MNER, which depends on noun phrases in sentences and general domain words to obtain visual cues. Besides, promptMNER \cite{prompt}, MRC-MNER \cite{mrc} and CAT-MNER \cite{cat} interpolate external knowledge, such as  machine reading comprehension, prompt and extended label words, to help multimodal understanding. However, the above studies still fail to capture implicit entity-object relations, due to insufficient NER supervision and specific entity names for cross-modal alignment. In this work, our method aims to enhance entity-object correlation by additional cross-modal generation.

\textbf{Cross-modal Generation.} Different from multimodal recognition tasks \cite{clip}, image caption \cite{hu2022scaling} and text-conditional image generation \cite{Hertz2022PrompttoPromptIE} are two typical cross-modal generation tasks for multimodal understanding. Chen et.al. \cite{Chen2022LearningDA} explore diverse description modes to produce controllable and informative image descriptions. Imagen Video \cite{Imagen} is not only capable of generating videos of high fidelity but also has a high degree of controllability and knowledge in various artistic styles. Inspired by cross-modal generation, Lat \cite{lat} introduces a latent cross-modal translation with global features to alleviate the information distortion for text-video retrieval. In this work, we focus on fine-grained cross-modal generation in latent space. However, directly applying previous generation methods on our Twitter datasets works poorly, due to the fact that iamge-text correlation from real-world social meadia is far worse than image caption datasets. To alleviate the semantic gap in our case, we first estimate the content that has the same meaning in two modalities, and then use such content for cross-modal generation with cycle-consistency loss \cite{cycle}.

\section{Our Method}
\subsection{Task Formulation}

Given a sentence $S=(w_1, w_2,...,w_n)$ comprised of $n$ words and a corresponding image $I$, MNER aims to assign an NER tag $y_i\in \mathcal{Y}$ for each word $w_i$, where $\mathcal{Y}$ is a pre-defined label set with standard BIO schema \cite{BIO}. The predefined entity types usually contain person (\texttt{PER}), location (\texttt{LOC}), organization (\texttt{ORG}), and miscellaneous (\texttt{MISC}).

\subsection{Overview}
As illustrated in Figure \ref{overview}, we present a novel bidirectional generative alignment method named BGA-MNER, consisting of a Transformer layer for each modality, $N$ bidirectional generative alignment layers (BGA layers) and a CRF decoder. The BGA layers in the visual branch and textual branch are designed similarly. Taking the textual branch as an example, Stage-refined Context Sampler module first extracts matched content from image-text pairs. Then, we feed the extracted textual part to text2image generator $G_{t2v}$ to synthesize the pseudo visual features. Such cross-modal generation is further ensured by cycle consistency. Finally, Textual Feature Extractor is a text-conditional hybrid attention layer 
 with real textual features and generated visual features.  

\textbf{Input Embedding.} Given a sentence $S$, we follow BERT \cite{bert} to tokenize it into a sequence of word embeddings. Then the special tokens [CLS] and [SEP] are inserted at the beginning and end positions of word embeddings. As a result, we feed $\mathbf{T}=\{T_i\}_{i=1}^{N_t} \in \mathbb{R}^{N_t \times d}$ with $N_t$ tokens to text branch as input.

        

To extract features from images, we leverage ViT-B/32 \cite{vit} from CLIP \cite{clip} as a visual feature extractor. Following \cite{hybrid}, the combination of image and salient objects according to the text is projected and flattened into patch embeddings $\mathbf{V}=\{V_i\}_{i=1}^{N_v} \in \mathbb{R}^{N_v \times d}$ with $N_v$ patches. We use the subscript $t$ and $v$ to represent textual and visual modality respectively. Since the design of the visual branch is almost identical to that of the text branch, we will elaborate our method on the textual branch for simplicity.



\subsection{Stage-refined Context Sampler}
Image and text normally contain irrelevant content which is unsuitable for generation. To boost the generation process, Stage-refined Context Sampler (SCS) is proposed to adaptively extract the matched content from two modalities. To avoid overwhelmed calculation, we design our SCS with multiple MLP layers to estimate the finer token/patch mask, which denotes keeping the content for generation, via considering the coarse estimation from previous layers. 

Taking textual SCS as an example, as illustrated in the bottom-right corner of Figure \ref{overview}, our SCS is designed in a recursive-refined fashion where the mask assignment depends on both current local features and previously selected global content. Concretely, In the $l$-th layer, current local feature $z^l_t \in \mathbb{R}^{N_t \times d}$ and previously selected global content $g^l_t \in \mathbb{R}^{1\times d}$ are obtained by two linear projectors: 

\begin{align}
\begin{aligned}
    z^l_t &= \mathrm{MLP_\phi}(T^l),\\
    g^l_t &= \mathrm{GAP}(\mathrm{MLP_\varphi}(T^l)* m_t^{l-1}),
\end{aligned}
\end{align}
where $\mathrm{GAP}$ denotes global average pooling and $m_t^{l-1}=\{m_{ti}^{l-1}\}_{i=1}^{N_t} \in \mathbb{R}^{N_t\times 1}$ is the token-wise textual mask decision from the previous layer. For the first layer, $m_t^{0}$ is initialized as an all-one vector. Then, the probability $p_t$ of retaining tokens is calculated by $\mathrm{Softmax}(\mathrm{MLP_\Phi}([z_t^l,g_t^l]))$,
where $[\cdot,\cdot]$ denotes the concatenation operation along channel dimension and $\phi, \varphi, \Phi$ represent different MLPs. We follow \cite{dynamicvit} to use GumbelSoftmax to sample binary decision mask $m_t^l$ from $p_t$ in a differentiable manner. In the visual branch, the mask decision $m_v^l$ for patch embeddings is obtained similarly. During the optimization procedure, the distributions of extracted two-modality content by SCS tend to be consistent under the generation supervision.

\subsection{Multi-level Cross-modal Generator}
Multi-level Cross-modal Generator (MCG) is to generate the pseudo content of one modality with the given sampled content of the other modality. It has a shared text2image generator $G_{t2v}$ and a shared image2text generator $G_{v2t}$ across layers to avoid overwhelmed parameters.

\textbf{Cross-modal Generator.}  We utilize transformer decoders with modality queries as the generator \cite{Zhang2019BERTScoreET,TransGAN}. Taking text2image generation in $l$-th layer for example, the pseudo visual embeddings $\hat{V}^l$ are calculated with the learnable visual query $Q_v \in \mathbb{R}^{N_v \times d}$ \cite{bert} and textual content $T^l $ with decision mask $m_t^l$:

\begin{equation}
 \hat{V}^l = G_{t2v}(T^l, Q_v, m_t^l).
\end{equation}

In detail, we reformulate the attention of transformer decoders in $G_{t2v}$ as\footnote{We omit the FeedForward Network and LayerNorm.}:
\begin{equation}
    \hat{V}^l = \operatorname{Softmax}(\frac{(Q_v W^Q)(T^l W^K)^T}{\sqrt{d}}+ \mathcal{M}) \cdot Q_v W^V,
\end{equation}
where $W^Q \in \mathbb{R}^{d \times d_q},W^K \in \mathbb{R}^{d \times d_k},W^V \in \mathbb{R}^{d \times d_v}$ are randomly initialized projection matrices. We set $d_q=d_k=d_v=d/h$ where $h$ is the number of heads in each multi-head attention layer. $\mathcal{M} \in \{0, -\text{inf}\}^{{N_v \times N_t}}$ is the attention mask. The element in $\mathcal{M}$ is set to 0 for keeping the unit and set to negative infinity for removing it. $\mathcal{M}$ is obtained by transposing $m_t^l$ and then broadcasting $(m_t^l -1)*\text{inf}$ from $1 \times N_t$ to $N_v \times N_t$ resolution.

\textbf{Reconstruction for Generation.} The reconstruction loss is used to ensure generating desired features. Specifically, the generated  textual $\hat{T}^l$ feature is calculated by:

\begin{equation}
    \hat{T}^l = G_{v2t}(V^l, Q_t, m_v^l),   
\end{equation}
where $Q_t \in \mathbb{R}^{N_v \times d}$ is the learnable textual query. Thus, the reconstruction loss for generation is:
\begin{equation}
        \mathcal{L}_{recon}^l \resizebox{.75\hsize}{!}{$=\sum\limits_{i=1}^{N_v} \mathcal{D}_{KL}(\hat{V}_i^l || V^l_i )\cdot m_{vi}^l + \sum\limits_{i=1}^{N_t} \mathcal{D}_{KL}(\hat{T}^l_i || T^l_i )\cdot m_{ti}^l,$}
\end{equation}
where $\mathcal{D}_{KL}$ denotes the Kullback-Leibler Divergence \cite{KL}.

\textbf{Cycle Consistency for Generation.} Apart from traditional reconstruction loss for cross-modal generation, we further take advantage of cycle consistency to ensure mutual latent translation. The intuition of cycle consistency is from CycleGAN \cite{cycle}: Translating a sentence from English to French and then translating it back from French to English, we should arrive at the original sentence. In multimodal generation, we expect the generated visual feature could be inversely generated to its original textual feature. We elaborate the detail of cycle generation as:
\begin{align}
\begin{aligned}
   & \bar{V}^l = G_{t2v}(\hat{T}^l, Q_v, \mathbbm{1}), \\
   & \bar{T}^l = G_{v2t}(\hat{V}^l, Q_t, \mathbbm{1}),
\end{aligned}
\end{align}
where $\bar{V}$ and $\bar{T}$ are cycle-generated visual and textual features. In cycle case, the content of $\hat{V}_l$ and $\hat{T}_l$ used for generation is mask-agnostic. Thus, we use all-one vector $\mathbbm{1}$. Finally, the cycle consistency loss in $l$-th layer is represented as:  
\begin{equation}
        \mathcal{L}_{cycle}^l \resizebox{.75\hsize}{!}{$=\sum\limits_{i=1}^{N_v} \mathcal{D}_{KL}(\bar{V}^l_i || V^l_i )\cdot m_{vi}^l + \sum\limits_{i=1}^{N_t} \mathcal{D}_{KL}(\bar{T}_i^l || T^l_i )\cdot m_{ti}^l.$}
\end{equation}

 A vital concern in applying cross-modal generation to MNER is the incomplete visual information. Previous studies use a uniform empty image as the alternative missing image \cite{umt}. But in generation task, such misalignment in image-text pair would lead to model collapse. To solve this issue, we exclude the generation loss ($\mathcal{L}_{recon}$ and $\mathcal{L}_{cycle}$) of incomplete text-image pairs. In other words, our BGA-MNER actively refuses the injection of noise from such invalid missing images, instead of only relying on gate function to remove the visual noise.

\subsection{Feature Extractor} We derive the formulation of feature extraction layer from the standard transformer layer. As shown in top-right corner of Figure \ref{main}, taking textual feature extractor as an example, the calculation of multi-head attention of textual feature extractor is performed on the hybrid key and value:


\begin{equation}
\resizebox{.85\hsize}{!}{$T^{l+1} = \operatorname{Softmax}(\frac{ (T^l W^Q)([T^l,\hat{V}^l] W^K)^T}{\sqrt{d}}) ([T^l, \hat{V}^l] W^V),$}
\end{equation}
where $[\cdot,\cdot]$ denotes the concatenation operation, $W^Q \in \mathbb{R}^{d \times d_q},W^K \in \mathbb{R}^{d \times d_k},W^V \in \mathbb{R}^{d \times d_v}$ are randomly initialized projection matrices. We set $d_q=d_k=d_v=d/h$ where $h$ is the number of heads in each multi-head attention layer.

The self-attention and cross-attention are assembled within a Transformer layer for (1) parameter-efficiency (2) reducing modality heterogeneity and (3) being compatible with BERT and ViT. During implementation, the textual feature extractor and visual feature extractor are initialized by loading the pretrained parameters of BERT layer and ViT layer, respectively.

\subsection{CRF Decoder}
After stacking $N$ bidirectional generative alignment layers, we apply Conditional Random Fields (CRF) \cite{HBiLSTM} decoder to perform the sequence labeling for MNER task. CRF considers the correlations between labels in neighborhoods and scores the whole sequence of labels. Given the final textual hidden state $T^N$, CRF decoder produces the probability of a predicted label sequence $y$ in:
\begin{equation}
    p_\theta(y \mid T^N)=\frac{\prod_{i=1}^n \psi\left(y_{i-1}, y_i, T^N\right)}{\sum_{y^{\prime} \in \mathcal{Y}} \prod_{i=1}^n \psi\left(y_{i-1}^{\prime}, y_i^{\prime}, T^N\right)},
\end{equation}
where $\theta$ denotes the parameters of the CRF decoder, $\psi\left(y_{i-1}, y_i, T^N\right)$ and $\psi\left(y_{i-1}^{\prime}, y_i^{\prime}, T^N\right)$ are the potential function. 
We use the maximum conditional likelihood estimation as the loss function for MNER:
\begin{equation}
   \mathcal{L}_{mner}= -\sum \log p_\theta(y\mid T^N).
\end{equation}

The overall loss is the combination of MNER loss and generation loss:
\begin{equation}
\mathcal{L}_{overall} = \mathcal{L}_{mner} + \alpha \cdot  \frac{1}{N}\sum_{l=1}^{N} (\mathcal{L}_{recon}^l + \mathcal{L}_{cycle}^l),
\end{equation}
where $\alpha$ is the coefficient factor for balancing the losses of two tasks.


\textbf{Inference.} Notably, different from interaction with real two-modality features, our method uses the generated visual feature to simulate the real one for cross-modal interaction. The real visual and textual features are not integrated directly. Therefore, it is unnecessary to process the visual branch during inference, which is another merit of our work. Thanks to the SCS that samples salient entity content in text, our MCG directly generates its corresponding visual features, rather than extracting potential visual clues from noisy images. 

\begin{table}[]\caption{The statistics of two multimodal Twitter datasets.}\label{dataset}
\centering
\begin{adjustbox}{width=0.9\columnwidth}
\begin{tabular}{l|ccc|ccc}
\toprule
    \multirow{2}*{Entity Type} & \multicolumn{3}{c|}{Twitter2015} & \multicolumn{3}{c}{Twitter2017} \\
    & Train & Dev & Test & Train & Dev & Test \\
\midrule
	Person           & 2217      & 552      & 1816     & 2943      & 626      & 621      \\
	Location         & 2091      & 522      & 1697     & 731       & 173      & 178      \\
	Organization     & 928       & 247      & 839      & 1674      & 375      & 395      \\
	Miscellaneous    & 940       & 225      & 726      & 701       & 150      & 157      \\
\midrule
	Total            & 6176      & 1546     & 5078     & 6049      & 1324     & 1351     \\
	Num of Twittets  & 4000      & 1000     & 3257     & 3373      & 723      & 723      \\
\bottomrule
\end{tabular}
\end{adjustbox}
\end{table}

\begin{table*}[!h]\caption{Performance comparison of different competitive text-based and multi-modal approaches on two Twitter datasets. AdaCAN-CNN-BiLSTM-CRF is abbreviated to A-C-BiLSTM-CRF.}
\centering
\begin{adjustbox}{width=2\columnwidth}
\begin{tabular}{c|cccc|ccc|cccc|ccc}
\toprule
    \multirow{3}*{Methods} & \multicolumn{7}{c|}{Twitter2015} & \multicolumn{7}{c}{Twitter2017} \\
\cmidrule{2-15}
    & \multicolumn{4}{c}{Single Type (F1)} & \multicolumn{3}{c|}{Overall} & \multicolumn{4}{c}{Single Type (F1)} & \multicolumn{3}{c}{Overall} \\
\cmidrule{2-15}
    & PER & LOC & ORG & MISC & P & R & F1 & PER & LOC & ORG & MISC & P & R & F1 \\
\midrule
    BiLSTM-CRF 		& 76.77 & 72.56 & 41.33 & 26.80 & 68.14 & 61.09 & 64.42  & 85.12 & 72.68 & 72.50 & 52.56 & 79.42 & 73.43 & 76.31 \\
	CNN-BiLSTM-CRF  & 80.86 & 75.39 & 47.77 & 32.61 & 66.24 & 68.09 & 67.15  & 87.99 & 77.44 & 74.02 & 60.82 & 80.00 & 78.76 & 79.37 \\
	HBiLSTM-CRF 	& 82.34 & 76.83 & 51.59 & 32.52 & 70.32 & 68.05 & 69.17  & 87.91 & 78.57 & 76.67 & 59.32 & 82.69 & 78.16 & 80.37 \\
	BERT            & 84.72 & 79.91 & 58.26 & 38.81 & 68.30 & 74.61 & 71.32  & 90.88 & 84.00 & 79.25 & 61.63 & 82.19 & 83.72 & 82.95 \\
	BERT-CRF        & 84.74 & 80.51 & 60.27 & 37.29 & 69.22 & 74.59 & 71.81  & 90.25 & 83.05 & 81.13 & 62.21 & 83.32 & 83.57 & 83.44 \\
\midrule
	A-C-BiLSTM-CRF  & 81.98 & 78.95 & 53.07 & 34.02 & 72.75 & 68.74 & 70.69  & 89.63 & 77.46 & 79.24 & 62.77 & 84.16 & 80.24 & 82.15 \\
	UMT 			& 85.24 & 81.58 & 63.03 & 39.45 & 71.67 & 75.23 & 73.41  & 91.56 & 84.73 & 82.24 & 70.10 & 85.28 & 85.34 & 85.31 \\
	FMIT            & 86.77 & \textbf{83.93} & \textbf{64.88} & 42.97 & 75.11 & \textbf{77.43} & 76.25  & 93.14 & \textbf{86.52} & 83.93 & 70.90 & 87.51 & 86.08 & 86.79 \\
	MRC-MNER        & 85.71 & 81.97 & 61.12 & 40.20 & 78.10 & 71.45 & 74.63  & 92.64 & 86.47 & 83.16 & \textbf{72.66} & \textbf{88.78} & 85.00 & 86.85 \\
        HVPNet   & - & - & - & - & 73.87 & 76.82 & 75.32 & - & - & - & - & 85.84 & 87.93 & 86.87 \\
	CAT-MNER        & 85.57 & 82.53 & 63.77 & \textbf{43.38} & 76.19 & 74.65 & 75.41  & 91.90 & 85.96 & 83.38 & 68.67 & 87.04 & 84.97 & 85.99 \\
	R-GCN           & 86.36 & 82.08 & 60.78 & 41.56 & 73.95 & 76.18 & 75.00  & 92.86 & 86.10 & 84.05 & 72.38 & 86.72 & 87.53 & 87.11 \\
	BGA-MNER(ours)  & \textbf{86.80} & 83.62 & 63.60 & 42.65 & \textbf{78.60} & 74.16 & \textbf{76.31}  & \textbf{93.71} & 85.55 & \textbf{85.71} & 71.05 & 87.71 & \textbf{87.71} & \textbf{87.71} \\
\bottomrule
\end{tabular}
\end{adjustbox}
\label{main}
\end{table*}

\section{Experiments}
\subsection{Datasets and Evaluation Metrics}
We test on two publicly benchmark Twitter datasets (Twitter2015 and Twitter2017) which are provided by \cite{twitter15} and \cite{twitter17}, respectively. Table \ref{dataset} shows the detail of two datasets, including the number of entities for each type and the size of train/dev/test data split. 


We use the Micro-F1 score (F1) of each type and overall precision (P), recall (R), and Micro-F1 score (F1) to evaluate the performance of the MNER models, which are widely used in many recent works \cite{cat,fmit}. In all experiments, we use the evaluation code provided by UMT \cite{umt} for fair comparison.

\subsection{Implementation Details}

Our method is implemented on one NVIDIA P100 GPU with Pytorch 1.7.0. We use the pre-trained uncased BERT-based model \cite{bert} as textual encoder, and ViT-B/32 from CLIP \cite{clip} as the visual encoder. Thus, we stack $N=11$ BGA layers and 1 Transformer layer. The maximum length of the sentence input is set to 128. The visual input includes the global image and objects detected by Faster-RCNN \cite{faster}. All optimizations are performed with the AdamW optimizer and a linear warmup of ratio 0.01. We set batch size to 16, learning rate to $3\times 10^{-5}$, and training epoch to 30. The coefficient factor $\alpha$ for balancing two-task loss is set to 0.001.

\subsection{Baselines}
We compare two groups of baseline models with our method. The first group is the representative text-based approaches for NER: (1) \textit{BiLSTM-CRF}, \textit{CNN-BiLSTM-CRF} \cite{CNN-BiLSTM-CRF} and \textit{HBiLSTM-CRF} \cite{HBiLSTM} are the NER models with bidirectional LSTM and CRF layer. The difference between them lies in the encoding layers to obtain character-level embedding. (2) \textit{BERT} \cite{bert} and \textit{BERT-CRF} exploit more powerful pretrained BERT compared with above methods.

The second group includes several MNER approaches. (1) \textit{AdaCAN-CNN-BiLSTM-CRF} \cite{twitter15} is a classical CNN+LSTM+CRF combination with an adaptive co-attention network to decide whether to attend to the image.  (2) \textit{UMT} \cite{umt} empowers Transformer with a multi-modal interaction module to capture the inter-modality dynamics. (3) \textit{MRC-MNER} \cite{mrc} and \textit{CAT-MNER} \cite{cat} use external knowledge to provide prior information about entity types and image regions. For fair comparison, we compare the results of CAT-MNER using uncased BERT-base as text encoder. (4) \textit{FMIT} \cite{fmit} is the state-of-the-art method using unified lattice structure and entity boundary detection for joint noun-object detection. (5) \textit{HVPNet} \cite{hvpnet} proposes a hierarchical visual prefix  to achieve more effective and robust performance. (6) \textit{R-GCN} \cite{gcn} constructs an inter-modal relation graph and an intra-modal relation graph to gather the image information most relevant to the current text and image from the dataset.

\subsection{Main Results}
Table \ref{main} shows the performance comparison of our method with baseline models on two benchmarks. In the text-based approaches, the powerful BERT and post-doc CRF could bring better performance. 
Apparently, the external knowledge of pretrained text encoder facilitates the understanding of complicated Twitter posts. For the CRF layer, it benefits models by sequential link constraint between two consecutive labels, which reduces the possibility of predicting unreasonable labels like assigning B-PER after I-PER. Besides, the multimodal approaches achieve 1.60\%-4.35\% improvement over the best text-based approaches. It demonstrates that the additional visual information is helpful for NER. 

We also compare our method with previous multimodal approaches. It is obvious that our method achieves state-of-the-art results on two datasets simultaneously. Particularly, our method obtains 76.31\% and 87.71\% overall F1 score on two datasets. Compared with previous best approaches, i,e., R-GCN and FMIT, our method also has great advantages. For example, even though only outperforming R-GCN with 0.6\% on Twitter2017, our method obtains 1.31\% improvement over it on Twitter2015. The significant improvement over attention-based approaches (UMT, R-GCN), visual grounding based approach (FMIT) and knowledge-based approaches (CAT-MNER, MRC-MNER) shows the superiority of our method over current approaches. 

\begin{table}[]\caption{Ablation study of each component on overall F1 score of two datasets. ``cycle'' denotes the cycle consistency loss.}
\centering
\label{ablation}
\begin{adjustbox}{width=0.85\columnwidth}
\begin{tabular}{c|c|c}
\toprule
    Settings & Twitter2015 & Twitter2017 \\
\midrule
    BGA-MNER & 76.31 & 87.71 \\
\midrule
	w/o SCS       & 75.88 ($\downarrow$0.43)  & 87.53($\downarrow$0.18)  \\
	w/o cycle     & 75.70 ($\downarrow$0.61)  & 87.32($\downarrow$0.39)  \\
	w/o SCS \& w/o cycle & 74.90 ($\downarrow$1.41)  & 86.96($\downarrow$0.75)  \\
	w/o SCS \& w/o MCG   & 71.81 ($\downarrow$4.50)   & 83.44($\downarrow$4.27)  \\
\bottomrule
\end{tabular}
\end{adjustbox}
\end{table}

\subsection{Discussion and Analysis}
\textbf{Ablation Study} To verify the effectiveness of each component, we report the ablation results in Table \ref{ablation}. First, after removing the stage-refined context sampler, the performance drops by 0.43\% and 0.18\%, respectively. It shows our SCS refines image-text alignment by removing irrelevant content. We further provide the visualization of SCS results in Figure \ref{scs}. It is easy to observe that our SCS extracts the matched content, including desired entities and matched nouns in image and text. In Figure \ref{scs} (a), our SCS simplifies the sentence to two names and a sport which are semantically consistent to the extracted image content. Such matched two-modality content is necessary for following cross-modal generation. Besides, removing the redundant content further boosts the generation process. As shown in Figure \ref{scs} (b), the word `Great day' is hard for generation. Removing such subjective descriptions which are common in social media can alleviate the semantic gap in understanding MNER examples. 

\begin{figure}[t]
\includegraphics[width=1\linewidth]{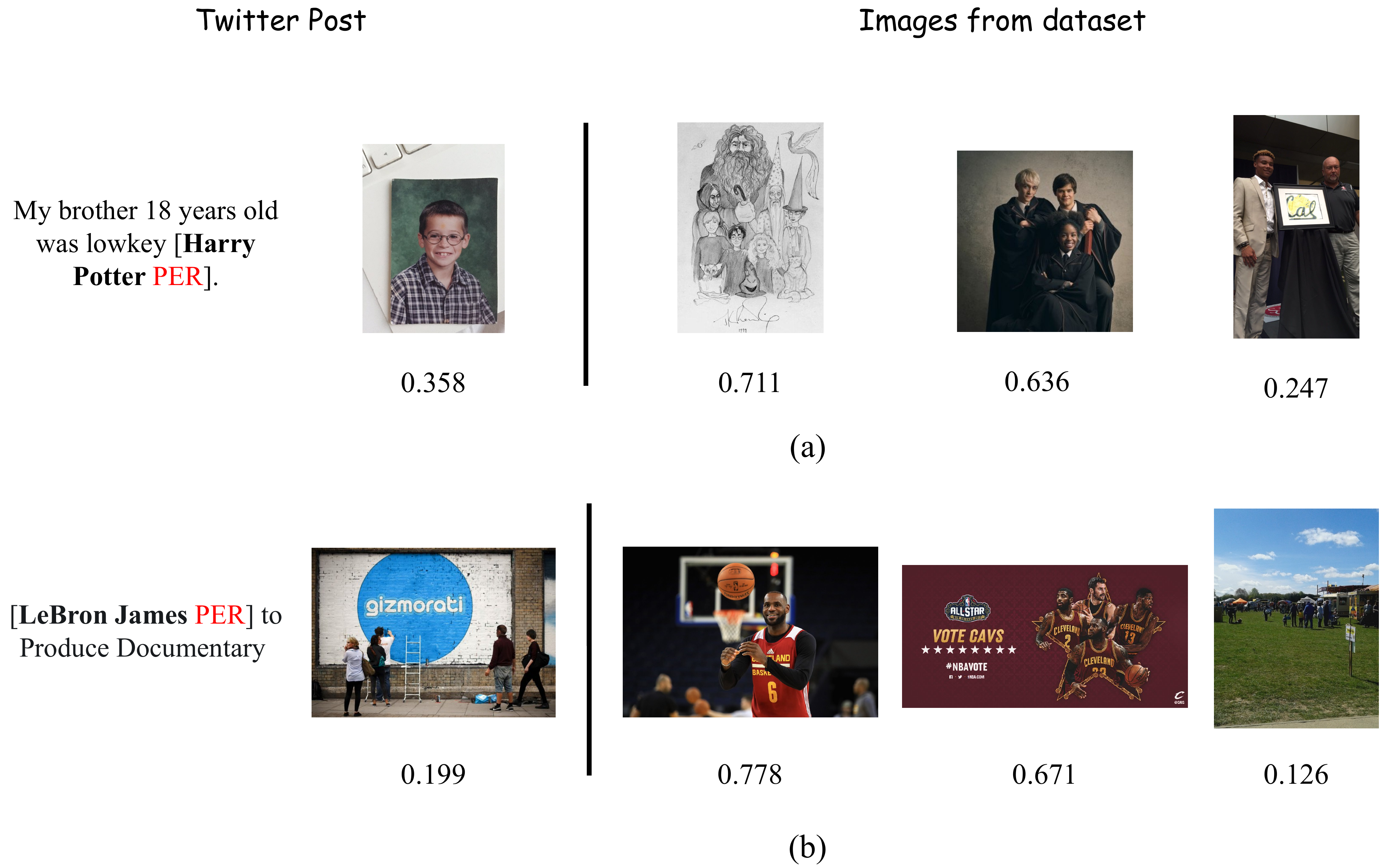}
\caption{Generative alignment analysis for implicit entity-object relations. We first generate the pseudo visual feature from MCG with extracted textual content produced by SCS. Then the similarity is calculated between pseudo visual feature, real visual feature from paired image (the first image in each row) and real visual feature from other images from dataset (the last three images).}
\label{relation}
\end{figure}

\begin{figure}[!h]
\centering
\includegraphics[width=1\linewidth]{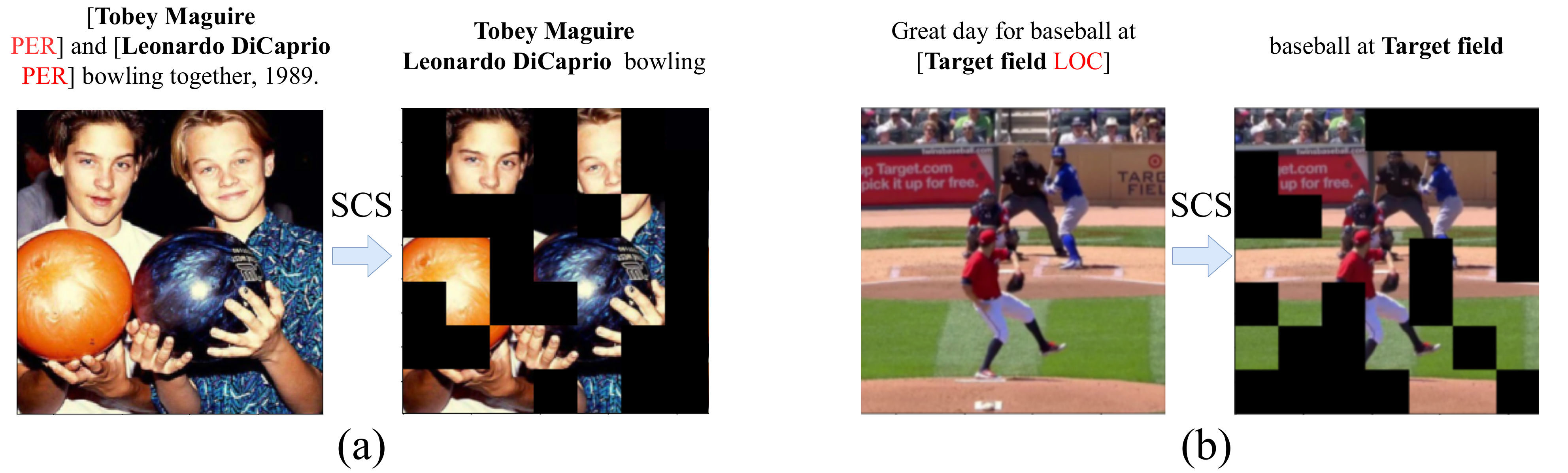}
\caption{Visualization of sampled content in image and text by SCS.}
\label{scs}
\end{figure}

As for the cycle consistency in MCG, it is proved to be helpful in enhancing multimodal understanding. Solely removing it decreases the performance by 0.61\% and 0.39\% while eliminating it with SCS deteriorates further. Finally, without SCS and MCG, our method degenerates to the text-based BERT-CRF, with 4.35\% and 4.27\% drops on two datasets. The significant improvement brought by MCG is from bidirectional generation with entity and its corresponding object, which introduce a direct and powerful constraint on entity-object alignment. To further investigate the effectiveness of our learned bidirectional generation, we conduct the following experiment. Given an image-text pair, we first generate the pseudo visual feature from MCG with extracted textual content produced by SCS 
Then the similarity is calculated between pseudo visual feature, real visual feature from paired image and real visual feature from other images from dataset. As shown the example in Figure \ref{relation} (a), the multisense entity `Harry Potter' needs additional information to classify, as it can stand for a movie, a book or a character. As we can see, the boy in the paired image is clearly not the character, which makes the entity semantically vague. In our method, we can obtain entity-related pseudo visual information by generation. The generated visual content is more related to the entity-related images (two images in the middle that contain character from Harry Potter story) than the irrelevant image (the forth image).

\begin{table}[t]\caption{Ablation study on using real or generated visual features for cross-modal interaction during inference.}
\centering
\begin{adjustbox}{width=1\columnwidth}
\begin{tabular}{c|cc}
\toprule
Features & Twitter2015 & Twitter2017 \\
\midrule
real textual + real visual     &   75.08          &    86.48         \\
 real textual + generated visual       &  76.31           &   87.71         \\
\bottomrule
\end{tabular}
\end{adjustbox}
\label{input}
\end{table}

\textbf{Image-free Inference.} A common concern raised naturally is why BGA-MNER does not use the real visual information during inference? The main reason is that our BGA-MNER can generate well-aligned pseudo visual features during inference, to avoid introducing potential noise in images. We first compare the results of using real and generated visual features in Table \ref{input}. Using real textual and generated visual features is about 1.2\% better than its counterpart. We believe the generated visual feature is more effective than the real one, especially for unmatched text-image pairs. As shown in Figure \ref{relation} (b), the entity `LeBorn James' does not exist in the paired image. However, we can still obtain the its pseudo visual information by generation where the generated visual content is more close to the images of the well-known player and discriminative with irrelevant images.

\begin{figure*}[!ht]
\includegraphics[width=00.9\linewidth]{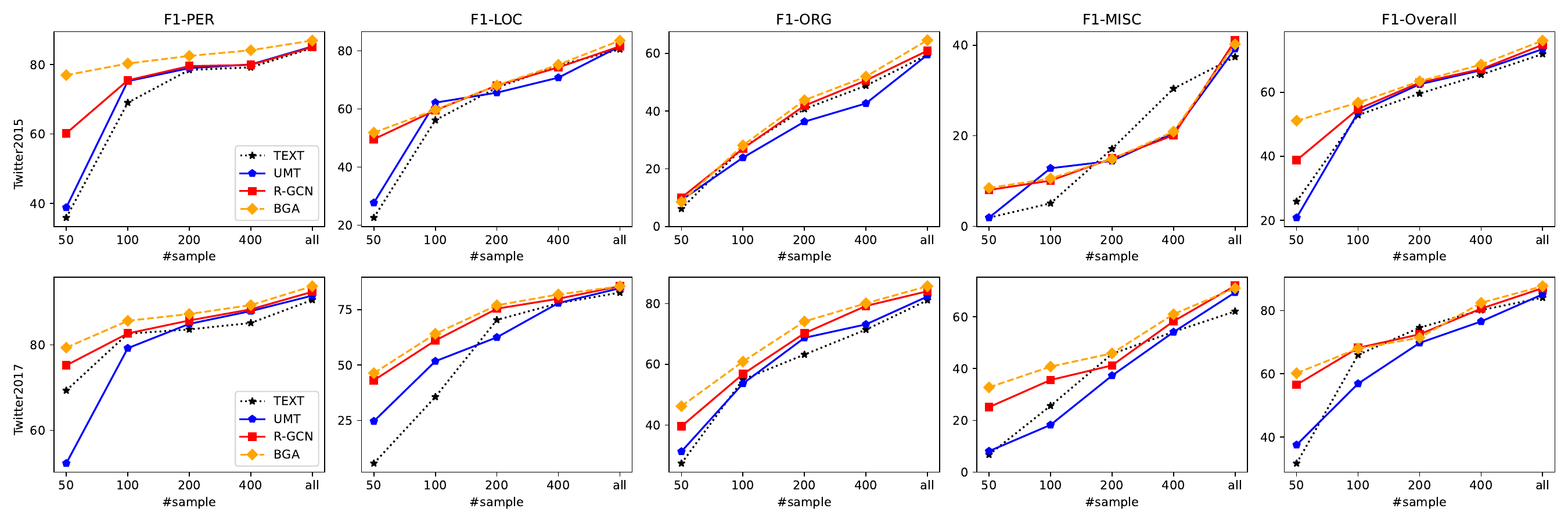}
\caption{Comparison in data efficiency evaluation on two datasets. TEXT denotes the text-based BERT-CRF while our BGA-MNER is abbreviated to BGA.}
\label{data}
\end{figure*}

\begin{table}[t]\caption{Ablation study on the visual encoder and model efficiency. All these methods use uncased BERT-base model as textual encoder. We report the total parameters of the whole network for  model efficiency analysis.}\label{backbone}
\centering
\begin{adjustbox}{width=1\columnwidth}
\begin{tabular}{c|c|c|cc}
\toprule
Method  & Visual Enc & Total Param  & Twitter2015 & Twitter2017 \\
\midrule
UMT     & ResNet152   & 206.3M&      73.41     &    85.31       \\
UMT     & ViT-B/32   & 233.6M&      73.52     &    85.64       \\
R-GCN  & ResNet152     & 172.2M&     75.00      &    87.11       \\
R-GCN  & ViT-B/32     &199.5M &   75.14        &     87.21      \\
CAT-MNER & ViT-B/32   & 198.5M&  75.41         &   85.99        \\
BGA-MNER & ViT-B/32   & 130.3M &  \textbf{76.31}         &   \textbf{87.71}       \\
\bottomrule
\end{tabular}
\end{adjustbox}
\end{table}

\textbf{Influence of Visual Encoder} Since most previous methods \cite{umt,gcn} adopt ResNet as visual encoder, we compare our method with them using the same encoder, \textit{i.e.,} ViT-b/32 from CLIP, for comparison fairness. As shown in Table \ref{backbone}, first, our BGA-MNER shows great superiority over CAT-MNER \cite{cat}, which originally uses the same visual encoder as our method. Second, we also fairly compare with other SOTA methods by replacing ResNet152 with ViT-B/32 from CLIP as the visual encoder in UMT and R-GCN. It is obvious that this encoder boosts their results with 0.1\%-0.3\% gain, however, our method still outperform with a large margin.

\textbf{Data and Model Efficiency Discussion.} For data efficiency, we randomly sample \{50, 100, 200, 400\} image-text pairs from the train split as the data-limited training set and then evaluate the model on the original test set. As shown in Figure \ref{data}, the line of BGA is generally higher than others in different conditions of samples, indicating our method is more data-effective than compared approaches. Specifically, in the extremely limited case with 50 examples, our BGA still outperforms R-GCN with nearly 20\% in F1-PER and 10\% in F1-Overall of Twitter2015. For the infrequent entity types, i.e., ORG in Twitter2015 and LOC, MISC in Twitter2017, our method also achieves better results than other works. 

For model efficiency, as shown in Table \ref{backbone}, our method is more lightweight than other methods. In detail, total textual SCS modules contain 11.5M parameters and text2image generator has 7.9M parameters. We can easily observe that (1) the additional parameters from SCS and MCG are negligible. (2) During inference we drop the whole visual branch, including SCS for image, image2text generation, and visual feature extractor. This design slims our method to a pure text-based BERT-CRF model.


\textbf{Cross-domain Generalization.} The difference in type distribution and data characteristics often brings significant performance gaps in practice. In Table \ref{cross}, we compare our method with other approaches in cross-domain scenarios. Cross-domain generalization analysis is implemented by training on the source dataset while testing on the target dataset. In all metrics, our BGA-MNER outperforms existing approaches by a large margin. Compared with UMT, our method outperforms it by 4.58\% and 4.04\% on F1 score. For CAT-MNER, our BGA-MNER shows superiority with 0.28\% and 0.33\% gain. These results demonstrate the strong generalization ability of our model. 


\begin{table}[]\caption{Comparison of the cross-domain generalization ability. Results are from \cite{cat,flat}.}
\centering
\begin{adjustbox}{width=0.9\columnwidth}
\begin{tabular}{c|ccc|ccc}
\toprule
    \multirow{2}*{Methods} & \multicolumn{3}{c}{Twitter2017$\rightarrow$Twitter2015} & \multicolumn{3}{|c}{Twitter2015$\rightarrow$Twitter2017} \\
    & P & R & F1 & P & R & F1 \\
\midrule
	UMT       & 64.67       & 63.59       & 64.13       & 67.80       & 55.23       & 60.87       \\
	UMGF      & 67.00       & 62.18       & 66.21       & 69.88       & 56.92       & 62.74       \\
	FMIT      & 66.72       & 69.73       & 68.19       & 70.65       & 59.22       & 64.43       \\
	CAT-MNER  & 74.86       & 63.01       & 68.43       & 70.69       & 59.44       & 64.58       \\
	BGA-MNER(ours)      & 72.17       & 67.98       & 68.71       & 70.81       & 59.60       & 64.91       \\
\bottomrule
\end{tabular}
\end{adjustbox}
\label{cross}
\end{table}

\textbf{Modality-missing Evaluation.} In social media, users do not always post with additional images. During inference, this modality-missing problem usually results in the failure of multimodal models trained on full-modality samples. We further analyze the property of existing MNER approaches against this issue in Table \ref{modality} by replacing all images in the test set with a uniform empty image. It can be observed that (1) UMT, MAF and R-GCN suffer from the unavailability of images with 0.2\%-0.6\% drops. These approaches rely on valid visual information for cross-modal interaction to recognize entities, leading to the performance decrease in modality-missing condition. (2) BGA-MNER and ITA \cite{ita} do not need the visual input, so as to be resistant to this issue. Moreover, our method outperforms ITA with 0.33\% and 2.22\% improvement on two datasets. We believe the multiple BGA layers facilitate the entity-object alignment by ensuring mutual translation at different semantic levels.

\begin{table}[]\caption{Modality-missing evaluation on overall F1 score of two datasets. The results are achieved by their official implementation.}\label{modality}
\centering
\begin{adjustbox}{width=0.7\columnwidth}
\begin{tabular}{c|ccccc}
\toprule
    Methods & Twitter2015 & Twitter2017 \\
\midrule
	UMT                   & 73.01($\downarrow$0.25) & 83.49($\downarrow$0.31) \\
	MAF                   & 73.06($\downarrow$0.41) & 85.79($\downarrow$0.22) \\
	R-GCN                 & 74.07($\downarrow$0.52) & 86.52($\downarrow$0.41) \\
	ITA                   & 75.98       & 85.49       \\
	BGA-MNER(ours)                  & 76.31       & 87.71       \\
\bottomrule
\end{tabular}
\end{adjustbox}
\end{table}

\begin{figure*}[t]
\includegraphics[width=0.75\linewidth]{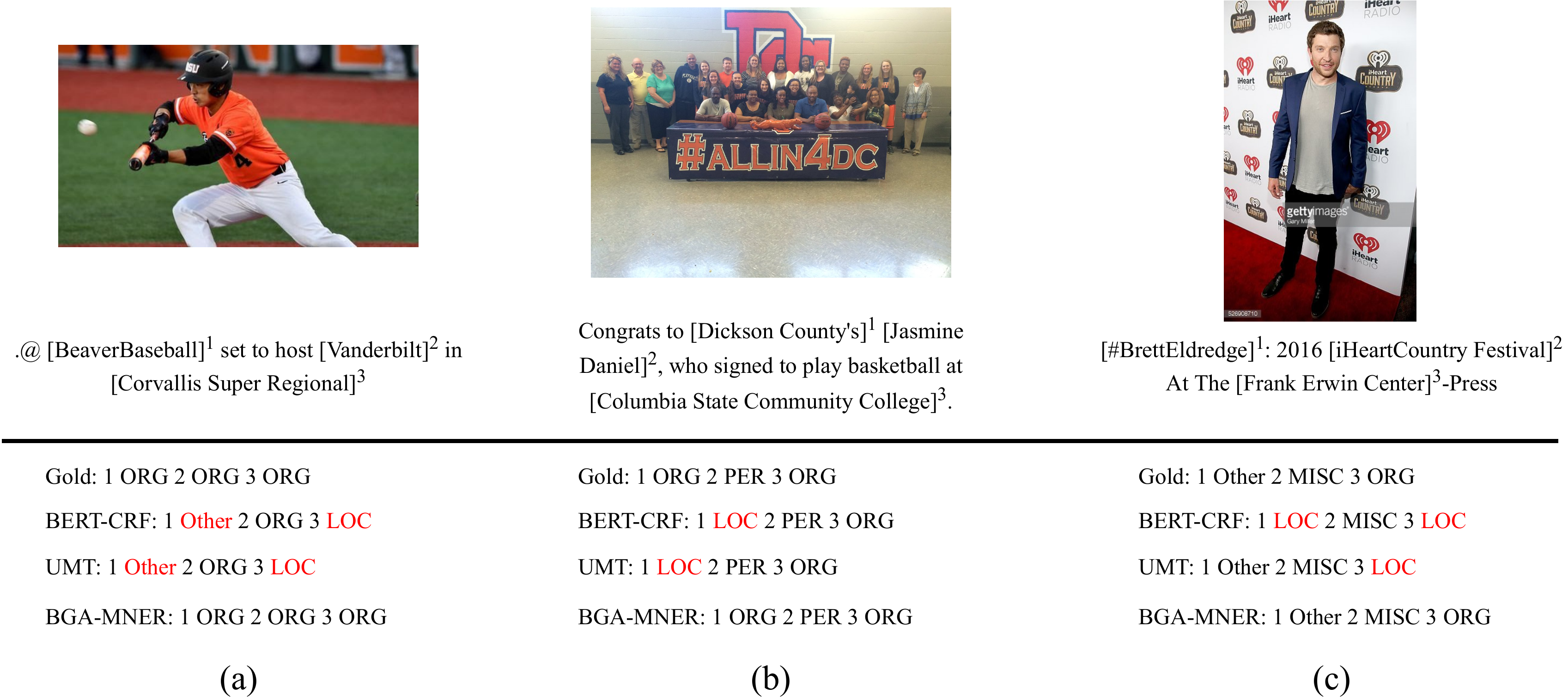}
\caption{Predictions of BERT-CRF, UMT and our BGA-MNER on three test samples.}
\label{case}
\end{figure*}


\textbf{Case Study} To better understand the advantage of our BGA-MNER, we select the representative samples with predictions of BERT-CRF \cite{bert} and UMT \cite{umt} in Figure \ref{case}. First, we can see that BERT-CRF and UMT heavily rely on nouns in text to recognize entities which may introduce harmful textual prior. For example, `County' and `Center' in case (b) and (c) mislead these two models with emphasis bias of location, leading to predict `Dickson County's' and `Frank Erwin Center' as LOC type. In contrary, our BGA-MNER takes advantage of bidirectional generative alignment to ensure mutual translation with additional generation supervision. We treat two modalities equally in bidirectional generation and highlight the content of each modality as input and output, alleviating the potential bias from text. Thus, our method could recover their true meaning as ORG type.

Furthermore, the understanding of multimodal context is superficial in UMT . The image of baseball sport in case (a) should be interpolated as event hosting and the personal photo in (c) actually describes an interview in a celebrating. However, the sport and person information simply understood by UMT is noise for MNER. In our BGA-MNER, we understand the pair of text and image in generative way. This manner focuses more on the content understanding with generation supervision, endowing our model with a powerful reasoning capability.

\section{Conclusion and Future Work}
In this paper, we propose a bidirectional generative alignment method named BGA-MNER. The main idea of our BGA-MNER is to use matched two-modality content for bidirectional generation, so as to provide direct and powerful constraints for entity-object cross-modal correlation. To be specific, we propose a new end-to-end Transformer MNER framework, which aligns the two modal features in a generative manner, leading to effectively capture the implicit entity-object relations. In our future work, we would like to apply our framework to vision-and-language pretraining as a more uniform structure than a conventional two-stream design. 

  \bibliographystyle{ACM-Reference-Format}
  \bibliography{sample-base}

\end{document}